\documentclass[10pt,twocolumn,letterpaper]{article}

\usepackage{cvpr}
\usepackage{times}
\usepackage{epsfig}
\usepackage{graphicx}
\usepackage{amsmath}
\usepackage{amssymb}
\usepackage{mathtools}
\usepackage{verbatim}
\usepackage{caption}
\usepackage{subcaption}
\usepackage{tabularx}
\usepackage{colortbl}
\usepackage{makecell}
\usepackage{etoolbox}
\usepackage{booktabs}
\usepackage{arydshln}
\usepackage{colortbl}
\usepackage{tabu}
\usepackage{multirow}
\usepackage{algorithm}
\usepackage[noend]{algpseudocode}
\usepackage{enumitem}
\usepackage[dvipsnames]{xcolor}
\usepackage[skip=1ex,font=small,labelsep=period]{caption}

\newcolumntype{Y}{>{\centering\arraybackslash}X}
\newcolumntype{R}{>{\raggedright\arraybackslash}X}
\newcolumntype{L}{>{\raggedleft\arraybackslash}X}

\DeclarePairedDelimiter{\ceil}{\lceil}{\rceil}
\newcommand{\Point}{\textbf{p}}
\newcommand{\trans}{\text{T}}

\definecolor{g1}{rgb}{1.00,1.00,1.00}
\definecolor{lightgray}{rgb}{0.9, 0.9, 0.9}
\definecolor{wincolor}{rgb}{0.95, 0.2, 0.2}

\newcommand{\win}[1]{\textcolor{wincolor}{\textbf{#1}}}
\newcommand{\second}[1]{\textcolor{NavyBlue}{\textbf{#1}}}
\newcommand{\dataset}[1]{{\fontfamily{cmtt}\selectfont #1}}

\makeatletter
\def\adl@drawiv#1#2#3{%
        \hskip.5\tabcolsep
        \xleaders#3{#2.5\@tempdimb #1{1}#2.5\@tempdimb}%
                #2\z@ plus1fil minus1fil\relax
        \hskip.5\tabcolsep}
\newcommand{\cdashlinelr}[1]{%
  \noalign{\vskip\aboverulesep
           \global\let\@dashdrawstore\adl@draw
           \global\let\adl@draw\adl@drawiv}
  \cdashline{#1}
  \noalign{\global\let\adl@draw\@dashdrawstore
           \vskip\belowrulesep}}
\makeatother

\makeatletter
\newlength{\qrr@dimen@}
\expandafter\pretocmd\csname tabular*\endcsname{\setlength{\qrr@dimen@}{#1}}{}{}
\newcommand*{\Rowcolor}[2][\tabcolsep]{%
    \ifx\relax#1\relax\else
        \kern-\the\dimexpr#1\relax
    \fi
    \makebox[0pt][l]{%
        \fboxsep=0pt
        \colorbox{#2}{%
            \strut\kern\qrr@dimen@
        }%
    }%
    \ifx\relax#1\relax\else
        \kern\the\dimexpr#1\relax
    \fi
    \ignorespaces
}
\makeatother
           
\newcommand\customparagraph[1]{\vspace{0.6em}\noindent\textbf{#1}}

\newcommand{\condexpectedvalue}[2]{E(#1 \, | \, #2)}

\usepackage[pagebackref=true,breaklinks=true,letterpaper=true,colorlinks,bookmarks=false]{hyperref}

\cvprfinalcopy 

\def\cvprPaperID{6311} 

\ifcvprfinal\fi
\begin{document}

\title{MAGSAC++, a fast, reliable and accurate robust estimator}

\author{Daniel Barath$^{12}$, Jana Noskova$^{1}$, Maksym Ivashechkin$^{1}$, and Jiri Matas$^{1}$\\
$^1$ Centre for Machine Perception, Department of Cybernetics \\
  Czech Technical University, Prague, Czech Republic \\
  $^2$ Machine Perception Research Laboratory, 
  MTA SZTAKI, Budapest, Hungary \\
    {\tt\small barath.daniel@sztaki.mta.hu}
}


\maketitle

\begin{abstract}
A new method for robust estimation, MAGSAC++\footnote{\url{https://github.com/danini/magsac}}, is proposed.
It introduces a new model quality (scoring) function that does not require the inlier-outlier decision, and a novel marginalization procedure formulated as an iteratively re-weighted least-squares approach.
We also propose a new sampler, Progressive NAPSAC, for RANSAC-like robust estimators. 
Exploiting the fact that nearby points often originate from the same model in real-world data, it finds local structures earlier than global samplers.
The progressive transition from local to global sampling does not suffer from the weaknesses of purely localized samplers.
On six publicly available real-world datasets for homography and fundamental matrix fitting,
MAGSAC++ produces results superior to the state-of-the-art robust methods. 
It is faster, more geometrically accurate and fails less often.

\end{abstract}

\section{Introduction}

\begin{figure}[t]
    \centering
	\begin{subfigure}[t]{0.99\columnwidth}
        \includegraphics[width=1.0\columnwidth]{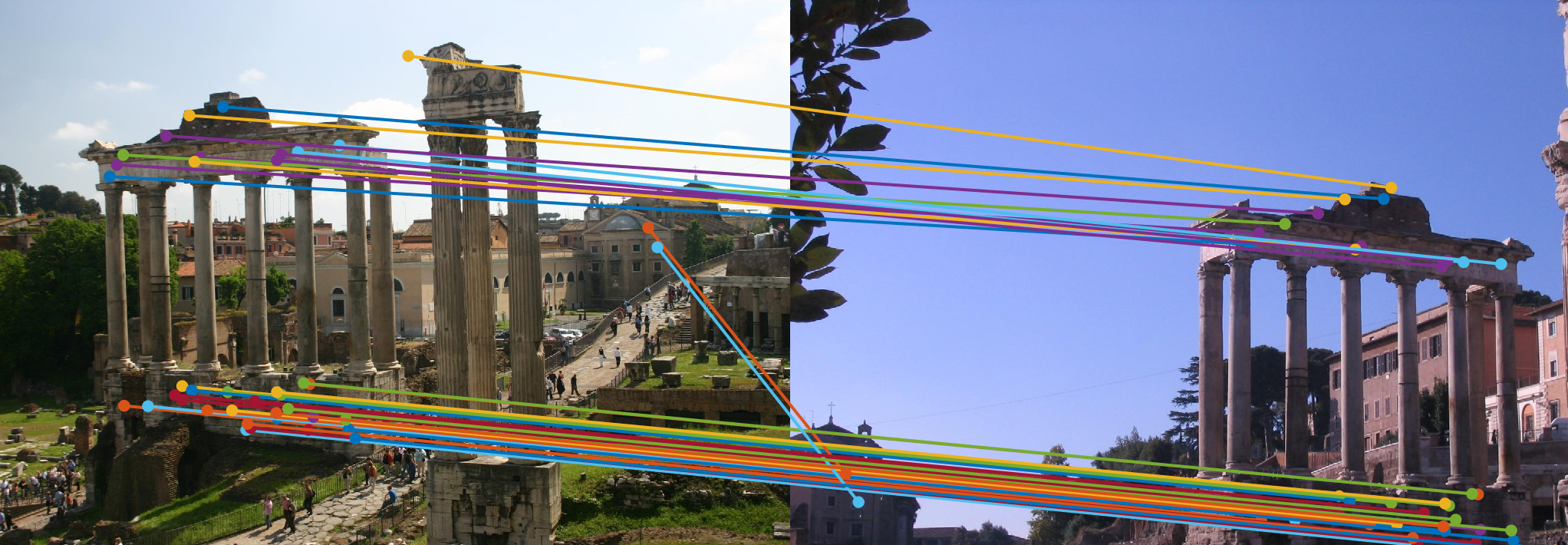}
        \caption{\dataset{Community Photo Collection dataset}~\cite{wilson2014robust}. }
	\end{subfigure}
	\begin{subfigure}[t]{0.99\columnwidth}
        \includegraphics[width=1.0\columnwidth]{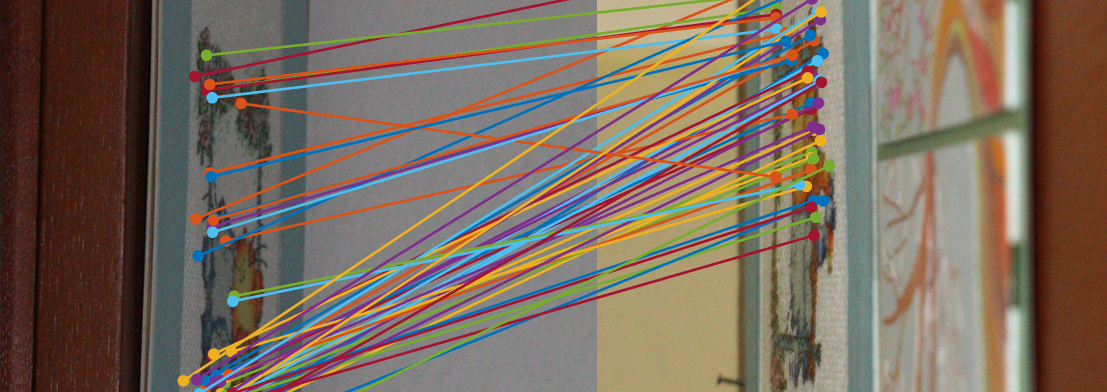}
        \caption{\dataset{ExtremeView dataset}~\cite{lebeda2012fixing}. }
	\end{subfigure}
	\begin{subfigure}[t]{0.99\columnwidth}
        \includegraphics[width=1.0\columnwidth]{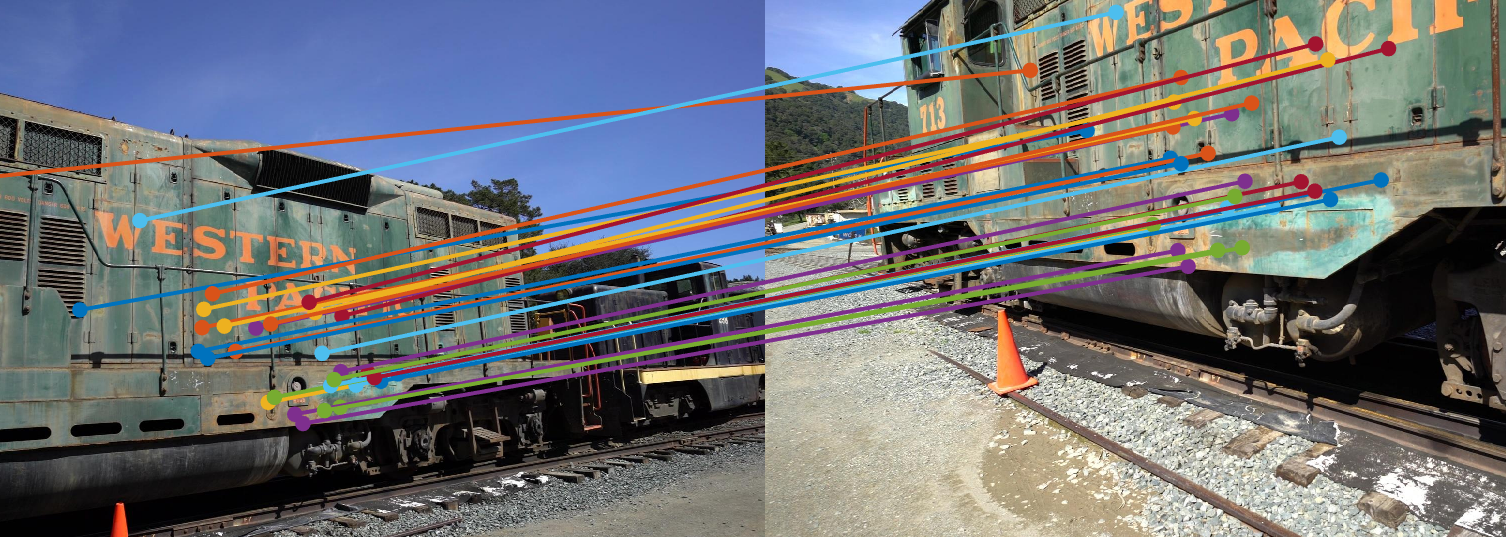}
        \caption{\dataset{Tanks and Temples dataset}~\cite{knapitsch2017tanks}.}
	\end{subfigure}
    \caption{Image pairs where all tested robust estimators (i.e., LMeDS~\cite{rousseeuw1984least}, RANSAC~\cite{fischler1981random}, MSAC~\cite{torr2000mlesac}, GC-RANSAC~\cite{barath2018graph}, MAGSAC~\cite{barath2019magsac}) failed, except the proposed MAGSAC++.
    Inlier correspondences found by MAGSAC++ are drawn by lines.}
    \label{fig:example_result}
\end{figure}


The RANSAC (RANdom SAmple Consensus) algorithm~\cite{fischler1981random} has become the most widely used robust estimator in computer vision. 
RANSAC and its variants have been successfully applied to a wide range of vision tasks, e.g., motion segmentation~\cite{torr1993outlier}, short baseline stereo~\cite{torr1993outlier,torr1998robust}, wide baseline matching~\cite{pritchett1998wide,matas2004robust,mishkin2015mods}, detection of geometric primitives~\cite{sminchisescu2005incremental}, image mosaicing~\cite{ghosh2016survey}, and to perform~\cite{zuliani2005multiransac} or initialize multi-model fitting~\cite{isack2012energy,pham2014interacting}.
In brief, RANSAC repeatedly selects minimal random subsets of the input point set and fits a model, e.g., a line to two 2D points or a fundamental matrix to seven 2D point correspondences.
Next, the quality of the estimated model is measured, for instance by the cardinality of its support, i.e., the number of inlier data points. Finally, the model with the highest quality, polished, e.g., by least squares fitting  of all inliers, is returned.


Since the publication of RANSAC, many modifications have been proposed improving the algorithm. 
For example, MLESAC~\cite{torr2000mlesac} estimates the model quality by a maximum likelihood process with all its beneficial properties, albeit under certain assumptions about data distributions. 
In practice, MLESAC results are often superior to the inlier counting of plain RANSAC, and they are less sensitive to the user-defined inlier-outlier threshold.
In MAPSAC~\cite{torr2002bayesian}, the robust estimation is formulated as a process that estimates both the parameters of the data distribution and the quality of the model in terms of maximum a posteriori. 




Methods for reducing the dependency on the inlier-outlier threshold include
MINPRAN~\cite{stewart1995minpran} which assumes that the outliers are uniformly distributed and finds the model where the inliers are least likely to have occurred randomly. 
Moisan et al.~\cite{moisan2012automatic} proposed a contrario RANSAC, selecting the most likely noise scale for each model.
Barath et al.~\cite{barath2019magsac} proposed the Marginalizing Sample Consensus method (MAGSAC) marginalizing over the noise $\sigma$ to eliminate the threshold from the model quality calculation.


The MAGSAC algorithm, besides not requiring a manually set threshold, was reported to be significantly more accurate than other robust estimators on various problems, on a number of datasets.
The improved accuracy originates from the new model quality function and $\sigma$-consensus model polishing.
The quality function marginalizes over the noise scale with the data interpreted as a mixture of uniformly distributed outliers and inliers with residuals having $\chi^2$-distribution.
The $\sigma$-consensus algorithm replaces the originally used least-squares (LS) fitting with weighted least-squares (WLS) where the weights are calculated via the marginalization procedure -- which requires a number of independent LS estimations on varying sets of points. 


Due to the several LS fittings, $\sigma$-consensus is slow. 
In~\cite{barath2019magsac}, a number of tricks (e.g., pre-emptive verification; down-sampling of $\sigma$ values) are proposed to achieve acceptable speed. 
However, \textit{MAGSAC is often significantly slower} than other robust estimators.
In this paper, we propose new quality and model polishing functions, reformulating the problem as an iteratively re-weighted least squares procedure. 
In each step, the weights are calculated making the same assumptions about the data distributions as in MAGSAC, but, without requiring a number of expensive LS fittings. 
The proposed MAGSAC++ and $\sigma$-consensus++ methods lead to \textit{more accurate results than the original MAGSAC algorithm, often, an order-of-magnitude faster}. 


In practice, there are also other ways of speeding up robust estimation.
NAPSAC~\cite{nasuto2002napsac} and PROSAC~\cite{chum2005matching} modify the RANSAC sampling strategy to increase the probability of selecting an all-inlier sample early. 
PROSAC exploits an a priori predicted inlier probability rank of the points and starts the sampling with the most promising ones.
PROSAC and other RANSAC-like samplers treat models without considering
that inlier points often are in the proximity of each other. 
This approach is effective when finding a global model with inliers sparsely distributed in the scene, for instance, the rigid motion induced by changing the viewpoint in two-view matching.
However, as it is often the case in real-world data, if the model is localized with inlier points close to each other, robust estimation can be significantly speeded by exploiting this in the sampling.


NAPSAC assumes that inliers are spatially coherent.
It draws samples from a hyper-sphere centered at the first, randomly selected, point. 
If this point is an inlier, the rest of the points sampled in its proximity are more likely to be inliers than the points outside the ball.
NAPSAC leads to fast, successful termination in many cases. 
However, it suffers from a number of issues in practice.
First, the models fit to local all-inlier samples are often imprecise due to the bad conditioning the points.
Second, in some cases, estimating a model from a localized sample leads to degenerate solutions.
For example, when fitting a fundamental matrix by the seven-point algorithm, the correspondences must originate from more than one plane.
Therefore, there is a trade-off between near, likely all-inlier, and global, well-conditioned, lower all-inlier probability samples.
Third, when the points are sparsely distributed and not spatially coherent, NAPSAC often fails to find the sought model. 


We propose in this paper, besides MAGSAC++, the Progressive NAPSAC (P-NAPSAC) sampler which merges the advantages of local and global sampling by drawing samples from gradually growing neighborhoods.
Considering that nearby points are more likely to originate from the same geometric model, P-NAPSAC finds local structures earlier than global samplers.
In addition, it does not suffer from the weaknesses of purely localized samplers due to progressively blending from local to global sampling, where the blending factor is a function of the input data. 

The proposed methods were tested on homography and fundamental matrix fitting on six publicly available real-world datasets. 
MAGSAC++ combined with P-NAPSAC sampler is superior to state-of-the-art robust estimators in terms of speed, accuracy and failure rate.
Example model estimations when all tested robust estimators, except MAGSAC++, failed, are shown in Fig.~\ref{fig:example_result}.

\section{MAGSAC++}

We propose a new quality function and model fitting procedure for MAGSAC~\cite{barath2019magsac}. 
It is shown that the new method can be formulated as an M-estimation solved by the iteratively reweighted least squares (IRLS) algorithm.

\customparagraph{The marginalizing sample consensus} (MAGSAC) algorithm is based on two assumptions.
\textit{First}, the noise level $\sigma$ is a random variable with density function $f(\sigma)$. 
Having no prior information, $\sigma$ is assumed to be uniformly distributed, $\sigma \sim  \mathcal{U}(0,\sigma_{\text{max}})$, where $\sigma_{\text{max}}$ is a user-defined maximum noise scale.
\textit{Second}, for a given $\sigma$, the residuals of the inliers are described by trimmed $\chi$-distribution\footnote{The square root of $\chi^2$-distribution.} with $n$ degrees of freedom  multiplied by $\sigma$ with density 
\begin{equation*}
   g(r \; | \; \sigma) = 2C(n)\sigma^{-n}\exp{(-r^2/2\sigma^2)}r^{n - 1},
\end{equation*}
for $r<\tau(\sigma)$ and   $ g(r \; | \; \sigma) =0 $ for $r\geq \tau(\sigma)$.
Constant $C(n) = (2^{n / 2}\Gamma(n / 2))^{-1}$ and, for $a > 0$, 
\begin{equation*}
    \Gamma (a)=\int_{0}^{+\infty} t^{a-1}  \exp {(-t)} {\mathrm d } t
\end{equation*}
is the gamma function, $n$ is the dimension of Euclidean space in which the residuals are calculated and $\tau(\sigma)$ is set to a high quantile (e.g., $0.99$) of the non-trimmed distribution.

Suppose that we are given input point set $\mathcal{P}$ and model $\theta$ estimated from a minimal sample of the data points as in RANSAC. 
Let $\theta_{\sigma} = F(I(\theta,\sigma,\mathcal{P}))$ be the model estimated from the inlier set $I(\theta,\sigma,\mathcal{P})$ selected using $\tau(\sigma)$ around the input model $\theta$. Scalar $\tau(\sigma)$ is the threshold which $\sigma$ implies; function $F$ estimates the model parameters from a set of data points; function $I$ returns the set of data points for which the point-to-model residuals are smaller than $\tau(\sigma)$.

For each possible $\sigma$ value, the likelihood of point $p \in \mathcal{P}$ being inlier is calculated as 
\begin{equation*}
    \text{P}(p \; | \; \theta_\sigma, \sigma) = 2 C(n) \sigma^{-n} D^{n - 1}(\theta_\sigma, p) \exp \left(\frac{-D^2(\theta_\sigma, p)}{2 \sigma^2} \right),
\end{equation*}
if $D(\theta_\sigma, p) \leq \tau(\sigma)$, where $D(\theta_\sigma, p)$ is the point-to-model residual. If $D(\theta_\sigma, p) > \tau(\sigma)$, likelihood $\text{P}(p \; | \; \theta_\sigma, \sigma)$ is $0$.

In MAGSAC, the final model parameters are calculated by weighted least-squares where the weights of the points come from marginalizing the likelihoods over $\sigma$.
It can be seen that, when marginalizing over $\sigma$, 
each $\text{P}(p \; | \; \theta_\sigma, \sigma)$ calculation requires to select the set of inliers and obtain $\theta_\sigma$ by LS fitting on them.
\textit{This step is time consuming} even with the number of speedups proposed in the paper. 

\customparagraph{In MAGSAC++}, we propose a new approach instead of the original one requiring several LS fittings when marginalizing over the noise level $\sigma$.
The proposed algorithm is an iteratively reweighted least squares (IRLS) where the model parameters in the $(i+1)$th step are calculated as follows:
\begin{equation} 
    \theta_{i+1}= \text{arg min}_{\theta} \sum_{p \in \mathcal{P}} w( D(\theta_i,p)) D^2(\theta,p), 
\end{equation} 
where the weight of point $p$ is 
\begin{equation} 
w( D(\theta_i,p))=\int \text{P}(p \; | \; \theta_i, \sigma)f(\sigma) {\mathrm d \sigma}   
\label{eq:weight_function}
\end{equation} 
and $\theta_0 = \theta$, i.e., the initial model from the minimal sample.

\subsection{Weight calculation}

It can be seen that the weight function defined in \eqref{eq:weight_function} is the marginal density of the inlier residuals as follows:
\begin{eqnarray*} 
w(r)=\int g ( r \; | \; \sigma) f(\sigma) {\mathrm d \sigma}.
\end{eqnarray*}
Let $\tau(\sigma) = k \sigma$ be the chosen quantile of the $\chi$-distribution. For $0\leq r\leq k \sigma_{\text{max}}$,
\begin{eqnarray*} 
w(r)= \frac{1}{\sigma_{\text{max}}} \int_{r/k}^{\sigma_{\text{max}}} g ( r|\sigma) {\mathrm d \sigma} = \\
\frac{1}{\sigma_{\text{max}}}C(n) 2^{ \frac{n-1}{2}} \left({\Gamma}\left( \frac{n-1}{2}, \frac{r^2}{2\sigma^2_{\text{max}}} \right) - {\Gamma}\left( \frac{n-1}{2}, \frac{k^2}{2} \right) \right)
\end{eqnarray*}
and, for $r > k \sigma_{\text{max}}$, $w(r) = 0$.  
%
%
Function 
\begin{equation*}
    \Gamma(a,x)=  \int_x^{+\infty}t^{a-1} \exp{(-t)}  {\mathrm dt}  
\end{equation*}
is the upper incomplete gamma function.

Weight $w(r)$ is positive and decreasing on interval $(0, \; k\sigma_{\text{max}})$. Thus there is a $\rho$-function of an M-estimator which is minimized by IRLS using $w(r)$ and each iteration guarantees a non-increase in its loss function (\cite{maronna2019robust}, chapter 9). Consequently, it converges to a local minimum.   
This IRLS with $\tau(\sigma) = 3.64 \sigma$, where $3.64$ is the $0.99$ quantile of $\chi$-distribution, will be called $\sigma$-consensus++. For problems using point correspondences, $n = 4$.
Parameter $\sigma_{\text{max}}$ is the same user-defined maximum noise level parameter as in MAGSAC, usually, set to a fairly high value, e.g., 10 pixels.
The $\sigma$-consensus++ algorithm is applied for fitting to a non-minimal sample and, also, as a post-processing to improve the output of any robust estimator. 

\subsection{Model quality function}

In order to be able to select the model interpreting the data the most, quality function $Q$ has to be defined. Let
\begin{equation} 
    Q(\theta, \mathcal{P}) = \frac{1}{L(\theta, \mathcal{P})},
    \label{eq:new_quality}
\end{equation}
where
\begin{equation*}
L(\theta, \mathcal{P})=  \sum_{p \in \mathcal{P}} \rho( D(\theta,p)),
\end{equation*}
is a loss function of the  M-estimator defined by our weight function $w(r)$.
Function $\rho(r) = \int_0^{r} xw(x) {\mathrm d x}$ for $r\in [0,+\infty)$.
For $0\leq r\leq k \sigma_{\text{max}}$,
%
%
%
\begin{eqnarray*}
\rho(r) = \frac{1}{\sigma_{\text{max}}}C(n) 2^{ \frac{n+1}{2}}
 [  \frac{\sigma^2_{\text{max}}}{2}  \gamma( \frac{n+1}{2}, \frac{r^2}{2\sigma^2_{\text{max}}}) + \\
 \frac{r^2}{4}(\Gamma ( \frac{n-1}{2}, \frac{r^2}{2\sigma^2_{\text{max}}})      
 - \Gamma( \frac{n-1}{2}, \frac{k^2}{2})) ]. 
\end{eqnarray*}
For $r> k \sigma_{\text{max}}$,
\begin{equation*}
    \rho(r) = \rho(k\sigma_{\text{max}})=  \sigma_{\text{max}}C(n) 2^{ \frac{n-1}{2}}
 \gamma( \frac{n+1}{2}, \frac{k^2}{2}),
\end{equation*}
where
\begin{equation*}
    \gamma(a,x)=  \int_0^{x}t^{a-1} \exp{(-t)}  {\mathrm dt}  
\end{equation*}
is the lower incomplete gamma function.
Weight $w(r)$ can be calculated precisely or approximately as in MAGSAC. However, the precise calculation can be done very fast by \textit{storing the values} of the complete and incomplete gamma functions \textit{in a lookup table}.
Then the weight and quality calculation becomes merely a few operations per point. 

MAGSAC++ algorithm uses \eqref{eq:new_quality} as quality function and $\sigma$-consensus++ for estimating the model parameters.

\section{Progressive NAPSAC sampling}

We propose a new sampling technique which gradually moves from local to global, assuming initially that localized minimal samples are more likely to be all-inlier.
If the assumption does not lead to termination, the process gradually moves towards the randomized sampling of  RANSAC. 

\subsection{N Adjacent Points SAmple Consensus}
The N Adjacent Points SAmple Consensus (NAPSAC) sampling technique~\cite{nasuto2002napsac} builds on the assumption that the points of a model are spatially structured and, thus, sampling from local neighborhoods increases the inlier ratio locally. In brief, the algorithm is as follows:\\[2mm]
\noindent
\textbf{1.}\hspace{3mm}Select an initial point $\Point_i$ randomly from all points. \\[1.5mm]
\noindent
\textbf{2.}\hspace{3mm}Find the set $\mathcal{S}_{i, r}$ of points lying within the hyper-sphere of radius $r$ centered at $\Point_i$.\\[1.5mm]
\noindent
\textbf{3.}\hspace{3mm}If the number of points in $\mathcal{S}_{i, r}$ is less than the minimal sample size then restart from step 1.\\[1.5mm]
\noindent
\textbf{4.}\hspace{3mm}Point $\Point_i$ and points from $\mathcal{S}_{i, r}$ selected uniformly form the minimal sample.\\[2mm]
There are three major issues of local sampling in practice. 
\textit{First}, it was observed that models fit to local all-inlier samples are often too imprecise (due to bad conditioning) for distinguishing all inliers in the data.
\textit{Second}, in some cases, estimating a model from a localized sample leads to degeneracy. 
For example, when fitting a fundamental matrix by the 7-point algorithm, the set of correspondences must originate from more than one plane.
This usually means that the correspondences are beneficial to be far. 
Therefore, purely localized sampling fails. 
\textit{Third}, in the case of having global structures, e.g., the rigid motion of the background in an image sequence, local sampling is much slower than global.
We, therefore, propose a transition between local and global sampling progressively blending from one into the other.

\begin{algorithm}
\begin{algorithmic}[1]
	\Statex{\hspace{-1.1em}\textbf{Input:} $\mathcal{P}$ -- points; $\mathcal{S}$ -- neighborhoods; $n$ -- point number}
    \State{$t_1, ..., t_n := 0$}\Comment{The hit numbers.}
    \State{$k_1, ..., k_n := m$}\Comment{The neighborhood sizes.}
   	\Statex{\hspace{-1.1em}\textbf{Repeat until termination:}}
    \Statex{\hrulefill}
   	\Statex{\textbf{Selection of the first point}:}
   	\State{Let $\Point_i$ be a random point.}\Comment{Selected by PROSAC.}
   	\State{$t_i := t_i + 1$}\Comment{Increase the hit number.}
   	\If{($t_i \geq T_{k_i}' \wedge k_i < n$)}
   	    \State{$k_i := k_i + 1$}\Comment{Enlarge the neighborhood.}
   	\EndIf
    \Statex{\hrulefill}
   	\Statex{\textbf{Semi-random sample $\mathcal{M}_{i,t_i}$ of size $m$}:}
   	\If{$\mathcal{S}_{i, k_i - 1} \neq \mathcal{P}$}
       	\State{Put $\Point_i$; the $k_i$th nearest neighbor; and $m - 2$ random}
        \Statex{\hskip\algorithmicindent  points from $\mathcal{S}_{i, k_i - 1}$ into sample $\mathcal{M}_{i,t_i}$.}
   	\Else 
   	    \State{Select $m - 1$ points from $\mathcal{P}$ at random.}
   	\EndIf
    \Statex{\hrulefill}
   	\Statex{\textbf{Increase the hit number of the points from $\mathcal{M}_{i,t_i}$}:}
   	\For{$\Point_j \in \mathcal{M}_{i,t_i} \setminus \Point_i$}\Comment{For all points in the sample,}
   	    \If {$\Point_i \in \mathcal{S}_{j, k_j}$}\Comment{if the $i$th one is close,}
   	        \State{$t_j := t_j + 1$}\Comment{increase the hit number.}
   	    \EndIf
   	\EndFor
   	\Statex{\textbf{Model parameter estimation}}
   	\State{Compute model parameters $\theta$ from sample $\mathcal{M}_{i,t_i}$.}
   	\Statex{\textbf{Model verification}}
   	\State{Find support, i.e., consistent data points, of the model with parameters $\theta$.} 
\end{algorithmic}
\caption{\bf Outline of Progressive NAPSAC. }
\label{alg:post_processing}
\end{algorithm}

\subsection{Progressive NAPSAC -- P-NAPSAC}

In this section, Progressive NAPSAC is proposed combining the strands of NAPSAC-like local sampling and the global sampling of RANSAC. 
The P-NAPSAC sampler proceeds as follows: the first, location-defining, point in the minimal sample is chosen using the PROSAC strategy. 
The remaining points, are selected {\it from a local neighbourhood}, according to their distances, applying a local PROSAC procedure. 
The process samples from the $m$ points nearest to the center defined by the first point in the minimal sample. 
The size of the local subset of points is increased data-dependently, as described below.
If no quality function is available, the first point is chosen at random similarly as in RANSAC, the other points are selected uniformly from a progressively growing neighbourhood.

In the case of having local models, the samples are more likely to contain inliers solely and, thus, trigger early termination. 
When the points of the sought model do not form spatially coherent structures, the gradual increment of neighborhoods leads to finding global structures not noticeably later than by using global samplers, e.g., PROSAC. 

\customparagraph{Growth function and sampling.}
The design of the growth function defining how fast the neighbourhood grows around a selected point $\Point_i$ must find the balance between the strict NAPSAC assumption -- entirely localized models -- and the RANSAC approach treating every model on a global scale. 
%
%

Let $ \{ \mathcal{M}_{i,j} \}_{j = 1}^{T(i)} = \{ \Point_{i}, \Point_{x_{i,j,1}}, ..., \Point_{x_{i, j, m - 1}} \}_{j = 1}^{T(i)}$ denote the sequence of samples $\mathcal{M}_{i,j} \subset \mathcal{P}^*$ containing point $\Point_i \in \mathcal{P}$ and drawn by some sampler (e.g., the uniform one as in RANSAC) where $m$ is the minimal sample size, 
$\mathcal{P}^*$ is the power set of $\mathcal{P}$, and $x_{i, j, 1}$, $...$, $x_{i, j, m - 1} \in \mathbb{N}^+$ are indices, referring to points in $\mathcal{P}$. In each $\mathcal{M}_{i,j} $, the points  are ordered with respect to their distances from $\Point_i$  and indices $j$ denote the order in which the samples were drawn.
%
%
%
The objective is to find a strategy which draws samples consisting of points close to the $i$th one and, 
then, samples which contain data points farther from $\Point_i$ are drawn progressively. 

Since the problem is quite similar to that of PROSAC, the same growth function can be used.
Let us define set $\mathcal{S}_{i,k}$
to be the smallest ball centered on $\Point_i$ and containing its $k$ nearest neighbours.
Let $T_k(i)$ be the number of samples from $\{ \mathcal{M}_{i,j} \}_{j = 1}^{T(i)}$ which contains $\Point_i$ and the other points are from $\mathcal{S}_{i,k}$. For the expected number of $T_k(i)$, holds: 
\begin{equation*}
    \condexpectedvalue{T_k(i)}{T(i)}  = T(i) \frac{\binom{k}{m - 1}}{\binom{n - 1}{m - 1}}  = T(i) \prod_{j = 0}^{m - 2} \frac{k - j}{n - 1 - j},
\end{equation*}
where $n$ is the number of data points. In this case, ratio $\condexpectedvalue{T_{k + 1}(i)}{T(i)} \, / \, \condexpectedvalue{T_{k}(i)}{T(i)}$ does not depend on $i$ and
$\condexpectedvalue{T_{k + 1}(i)}{T(i)}$ can be recursively defined as
\begin{equation*}
    \condexpectedvalue{T_{k + 1}(i)}{T(i)} = \frac{k + 1}{k + 2 - m} \condexpectedvalue{T_k(i)}{T(i)}.
\end{equation*}
We approximate it by integer function  $T_{k + 1}' = T_k' + \ceil{\condexpectedvalue{T_{k + 1}(i)}{T(i)} - \condexpectedvalue{T_{k}(i)}{T(i)}}$, where $T_1' = 1$ for all $i$. Thus, $T_k'$ is $i$-independent.
Growth function $g(t) = \min \{ k \, : \, T_k' \geq t \}$, i.e., for integer $t$, $g(t) = k$ where $T_k' = t$.
Let $t_i$ be the number of samples including $\Point_i$. For set $\mathcal{S}_{i,g(t_i)}$,
$t_i$ is approx.\ the mean number of samples drawn from $\mathcal{S}_{i,g(t_i)}$ if
the random sampler of RANSAC is used.

In the proposed P-NAPSAC sampler, the neighbourhood $\mathcal{S}_{i,k}$ of $\Point_i$ grows if 
$g(t_i)=k$, i.e., the number of drawn samples containing the $i$th point is approximately equal to the mean number of the samples drawn from this neighbourhood by the random sampler.

The $t_i$th sample $\mathcal{M}_{i, t_i}$, containing $\Point_i$, is $\mathcal{M}_{i,t_i} = \{ \Point_{i}, \Point^{*}(g(t_i)) \} \cup \mathcal{M}_{i,t_i}'$, %
where $\mathcal{M}_{i,t_i}' \subset \mathcal{S}_{i, g(t_i) - 1}$ is a set of $|\mathcal{M}_{i,t_i}'| = m - 2$ data points, excluding $\Point_i$ and $\Point^{*}(g(t_i))$, randomly drawn from $\mathcal{S}_{i, g(t_i) - 1}$. 
Point $\Point^{*}(g(t_i))$ is the $g(t_i)$-th nearest neighbour of point $\Point_i$.

\textit{Growth of the hit number.} Given point $\Point_i$, the corresponding $t_i$ is increased in two cases. 
First, $t_i \leftarrow t_i + 1$ when $\Point_i$ is selected to be the center of the hyper-sphere.
Second, $t_i$ is increased when $\Point_l$ is selected, the neighborhood of $\Point_l$ contains $\Point_i$ and, also, that of $\Point_i$ contains $\Point_l$. 
Formally, let $\Point_l$ be selected as the center of the sphere ($l \neq i \; \wedge \; l \in [1, n]$). 
Let sample $\mathcal{M}_{l,j} = \{ \Point_l, \Point_{x_{l, j, 1}}, ..., \Point_{x_{l, j, m - 1}}  \}$ be selected randomly as the sample in the previously described way. If $i \in \{ x_{l, j, 1}, ..., x_{l, j, m - 1} \}$ (or equivalently, $\Point_i \in \mathcal{M}_{l,j}$) and $\Point_l \in \mathcal{S}_{i,g(t_i)}$ then $t_i$ is increased by one. 

The sampler (see Alg.~\ref{alg:post_processing}) can be imagined as a PROSAC sampling defined for every $i$th point independently, where the sequence of samples for the $i$th point depends on its neighbors. 
After the initialization, the first main step is to select $\Point_i$ as the center of the sphere and update the corresponding $t_i$.
Then a semi-random sample is drawn consisting of the selected $\Point_i$, its $k_i$th nearest neighbour and $m - 2$ random points from $\mathcal{S}_{i, k_i - 1}$ (i.e., the points in the sphere around $\Point_i$ excluding the farthest one).
Based on the random sample, the corresponding $t$ values are updated. 
Finally, the implied model is estimated, and its quality is measured.

\customparagraph{Relaxation of the termination criterion.} 
We observed that, in practice, the termination criterion of RANSAC is conservative and not suitable for finding local structures early. 
The number of required iterations $r$ of RANSAC is  
\begin{equation}
    r = \log(1 - \mu) / \log(1 - \eta^{m}),
    \label{eq:ransac_iterations}
\end{equation}
where $m$ is the size of a minimal sample, $\mu$ is the required confidence in the results and $\eta$ is the inlier ratio.
This criterion does not assume that the points of the sought model are spatially coherent, i.e., the probability of selecting a all-inlier sample is higher than $\eta^m$. 
Local structures typically have low inlier ratio. 
Thus, in the case of low inlier ratio, Eq.~\ref{eq:ransac_iterations} leads to too many iterations even if the model is localized and is found early due to the localized sampling.

\begin{figure}[t]
    \centering
	\begin{subfigure}[t]{0.49\columnwidth}
	    \includegraphics[width=1.0\columnwidth]{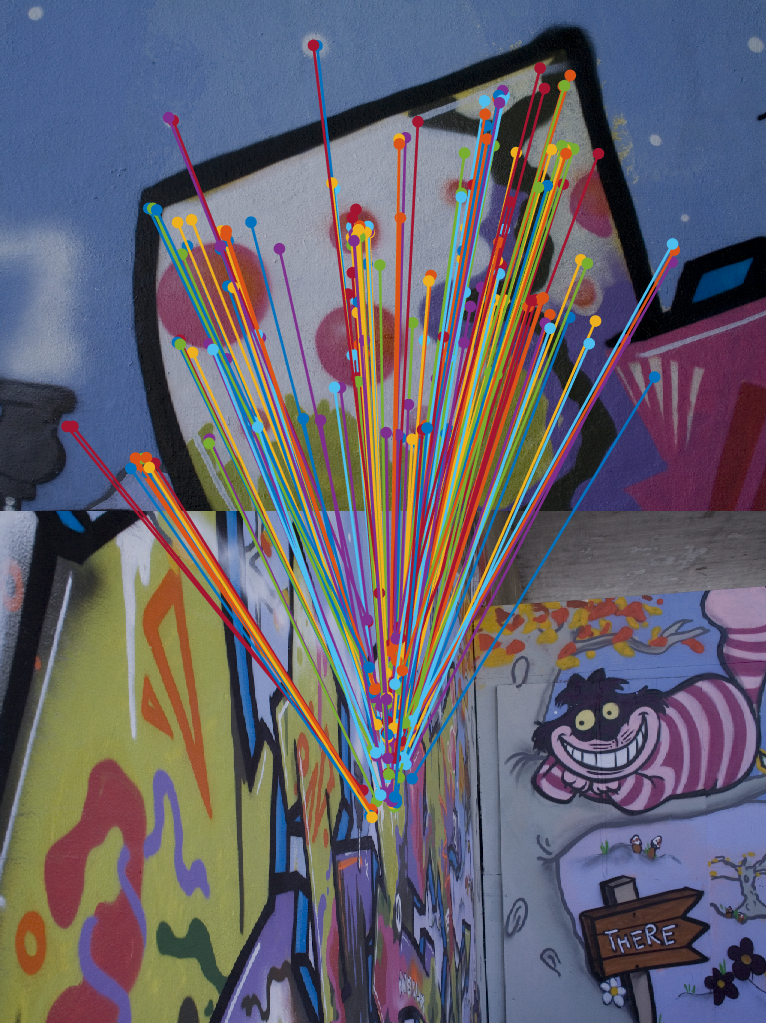}
	    \caption{
	    P-NAPSAC made $18\,302$ iterations in $0.49$ secs. PROSAC made $84\,831$ in $1.76$ secs. Scene "There".}
	\end{subfigure}\hfill
	\begin{subfigure}[t]{0.49\columnwidth}
	    \includegraphics[width=1.0\columnwidth]{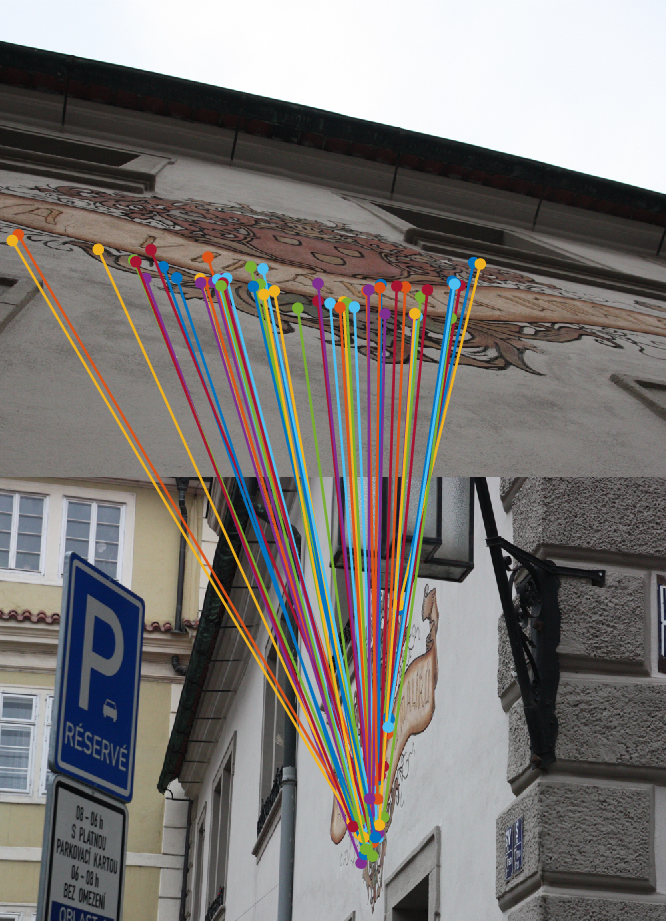}
	    \caption{P-NAPSAC made $65\,842$ iterations in $0.84$ secs. PROSAC made $99\,913$ in $1.28$ secs. Scene "Vin". }
	\end{subfigure}\hfill
    \caption{Example image pairs from the \dataset{EVD} dataset for homography estimation.
    Inlier correspondences are marked by a line segment joining the corresponding points.}
    \label{fig:example_result_2}
\end{figure}

A simple way of terminating earlier is to relax the termination criterion.
It can be easily seen that the number of iterations $r'$ for finding a model with $\eta + \gamma$ inlier ratio is 
\begin{equation}
    r' = \log(1 - \mu) / \log(1 - (\eta + \gamma)^{m}),
    \label{eq:relaxed_ransac_iterations}
\end{equation}
where $\gamma \in [0, 1 - \eta]$ is a relaxation parameter.

\customparagraph{Fast neighbourhood calculation.}
Determining the spatial relations of all points is a time consuming operation even by applying approximating algorithms, e.g., the Fast Approximated Nearest Neighbors method~\cite{muja2009fast}. 
In the sampling of RANSAC-like methods, the primary objective is to find the best sample early and, thus, spending significant time initializing the sampler is not affordable.
Therefore, we propose a multi-layer grid for the neighborhood estimation which we describe for point correspondences.
It can be straightforwardly modified considering different input data.

Suppose that we are given two images of size $w_l \times h_l$ ($l \in \{ 1,2 \}$) and a set of point correspondences $\{ (\Point_{i,1}, \Point_{i,2}) \}_{i = 1}^n$, where $\Point_{i,l} = [u_{i,l}, \; v_{i,l}]^\trans$.
A 2D point correspondence can be considered as a point in a four-dimensional space. 
Therefore, the size of a cell in a four-dimensional grid $\mathcal{G}_\delta$ constrained by the sizes of the input image is $\frac{w_1}{\delta} \times \frac{h_1}{\delta} \times \frac{w_2}{\delta} \times \frac{h_2}{\delta}$, where $\delta$ is parameter determining the number of divisions along an axis. 
Function $\Sigma(\mathcal{G}_\delta, [u_{i,1}, \; v_{i,1} \; u_{i,2}, \; v_{i,2}]^\trans)$ returns the set of correspondences which are in the same 4D cell as the $i$th one. Thus, $|\Sigma(\mathcal{G}_\delta, ...)|$ is the cardinality of the neighborhood of a particular point.
Having multiple layers means that we are given a sequence of $\delta$s such that: $\delta_1 > \delta_2 > ... \ > \delta_d \geq 1$. For each $\delta$, the corresponding $\mathcal{G}_{\delta_k}$ grid is constructed.
For the $i$th correspondence during its $t_i$th selection, the finest layer $\mathcal{G}_{\delta_{\max}}$ is selected which has enough points in the cell in which $\Point_i$ is stored. 
Parameter $\delta_{\max}$ is calculated as $\delta_{\max} := \max \{ \delta_k \; : \; k \in [1, d] \wedge |\mathcal{S}_{i, g(t_i) - 1}| \leq |\Sigma(\mathcal{G}_{\delta_k}, ...)| \}$.

In P-NAPSAC, $d = 5$, $\delta_1 = 16$, $\delta_2 = 8$, $\delta_3 = 4$, $\delta_4 = 2$ and $\delta_5 = 1$. When using hash-maps and an appropriate hashing function, the implied computational complexity of the grid creation is $\mathcal{O}(n)$. For the search, it is $\mathcal{O}(1)$. Note that $\delta_5 = 1$ leads to a grid with a single cell and, therefore, does not require computation. 



\section{Experimental Results}

In this section, we evaluate the accuracy and speed of the two proposed algorithms.
First, we test MAGSAC++ on fundamental matrix and homography fitting on six publicly available real-world datasets.
Second, we show that Progressive NAPSAC sampling leads to faster robust estimation than the state-of-the-art samplers. 
Note that \textit{these contributions are orthogonal} and, therefore, can be used together to achieve high performance efficiently -- by using MAGSAC++ with P-NAPSAC sampler.

\subsection{Evaluating MAGSAC++}

\customparagraph{Fundamental matrix} estimation was evaluated on the benchmark of~\cite{bian2019evaluation}.
The \cite{bian2019evaluation} benchmark includes: (1) the \dataset{TUM} dataset~\cite{sturm2012benchmark} consisting of videos of indoor scenes. Each video is of resolution $640 \times 480$.
(2) The \dataset{KITTI} dataset~\cite{geiger2012we} consists of consecutive frames of a camera mounted to a moving vehicle. The images are of resolution $1226 \times 370$.
Both in \dataset{KITTI} and \dataset{TUM}, the image pairs are short-baseline.
(3) The \dataset{Tanks and Temples} (\dataset{T\&T}) dataset~\cite{knapitsch2017tanks} provides images of real-world objects for image-based reconstruction and, thus, contains mostly wide-baseline pairs. 
The images are of size from $1080 \times 1920$ up to $1080 \times 2048$.
(4) The \dataset{Community Photo Collection} (\dataset{CPC}) dataset~\cite{wilson2014robust} contains images of various sizes of landmarks collected from Flickr. 
In the benchmark, $1\,000$ image pairs are selected randomly from each dataset. SIFT~\cite{lowe1999object}
correspondences are detected, filtered by the standard SNN ratio test~\cite{lowe1999object} and, finally, used for estimating the epipolar geometry.

The compared methods are RANSAC~\cite{fischler1981random}, LMedS~\cite{rousseeuw1984least}, MSAC~\cite{torr2000mlesac}, GC-RANSAC~\cite{barath2018graph}, MAGSAC~\cite{barath2019magsac}, and MAGSAC++.
All methods used P-NAPSAC sampling, pre-emptive model validation and degeneracy testing as proposed in USAC~\cite{raguram2013usac}. 
The confidence was set to $0.99$. 
For each method and problem, we chose the threshold maximizing the accuracy.
For homography fitting, it is as follows: MSAC and GC-RANSAC ($5.0$ pixels); RANSAC ($3.0$ pixels); MAGSAC and MAGSAC++ ($\sigma_{\text{max}}$ was set considering $50.0$ pixels threshold).
For fundamental matrix fitting, it is as follows: RANSAC, MSAC and GC-RANSAC ($0.75$ pixels); MAGSAC and MAGSAC++ ($\sigma_{\text{max}}$ which $5.0$ pixels threshold implies).
The used error metric is the symmetric geometric
distance~\cite{zhang1998determining} (SGD) which compares two fundamental matrices by iteratively generating points on the borders of the images and, then, measuring their epipolar distances. 
All methods were in C++.

\begin{figure}[t]
    \centering
    \includegraphics[width=0.495\columnwidth]{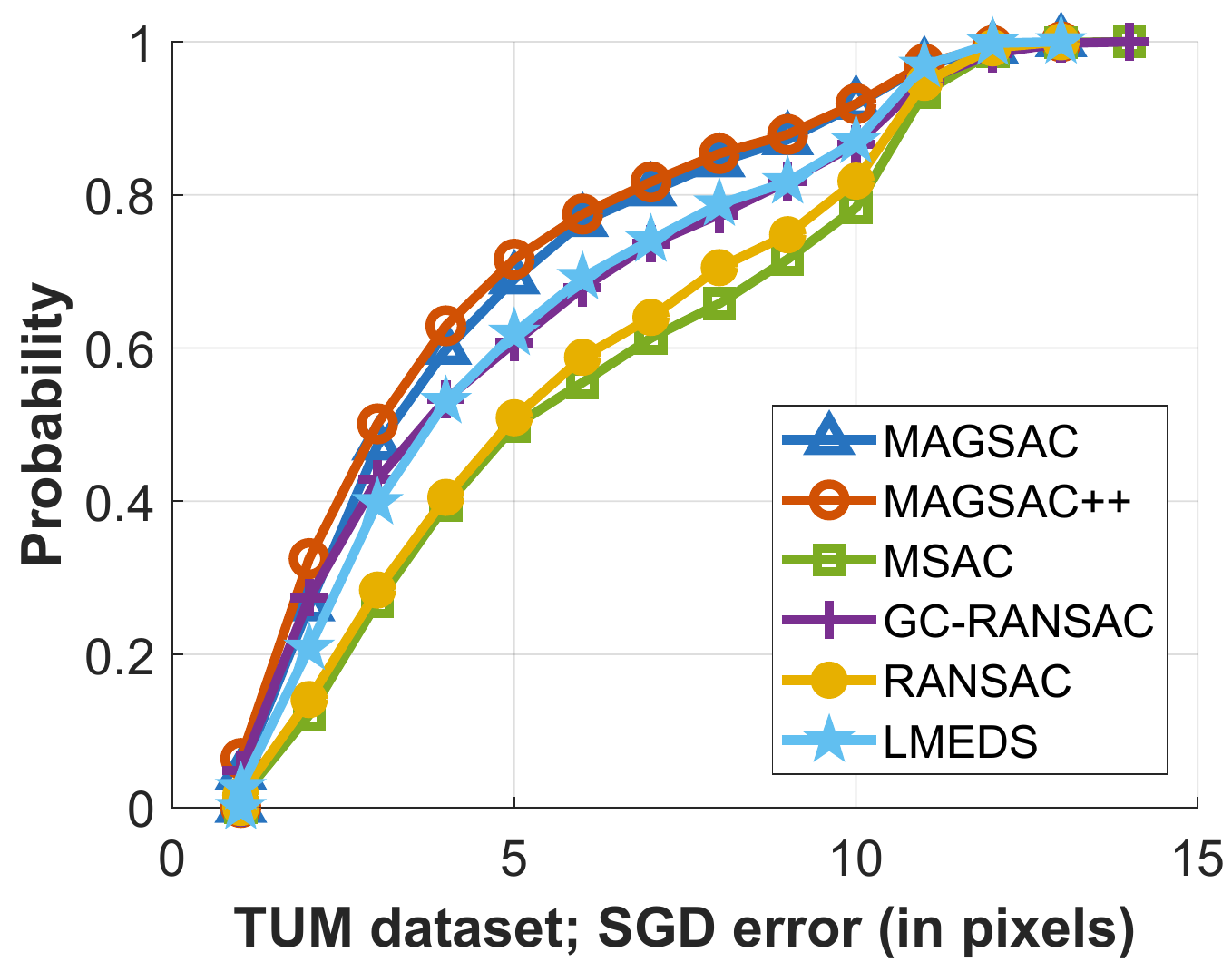}\hfill
    \includegraphics[width=0.495\columnwidth]{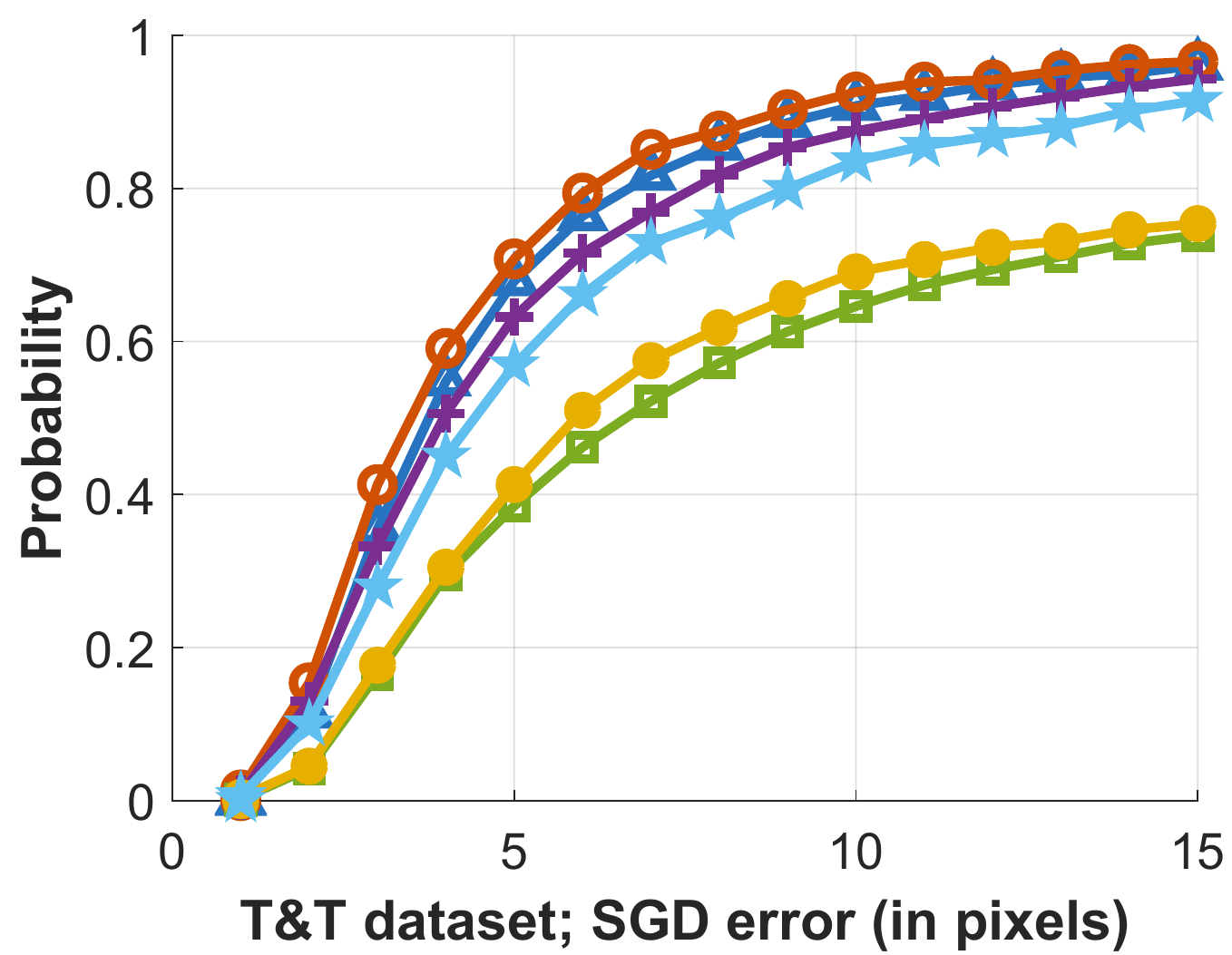}\\
    \includegraphics[width=0.495\columnwidth]{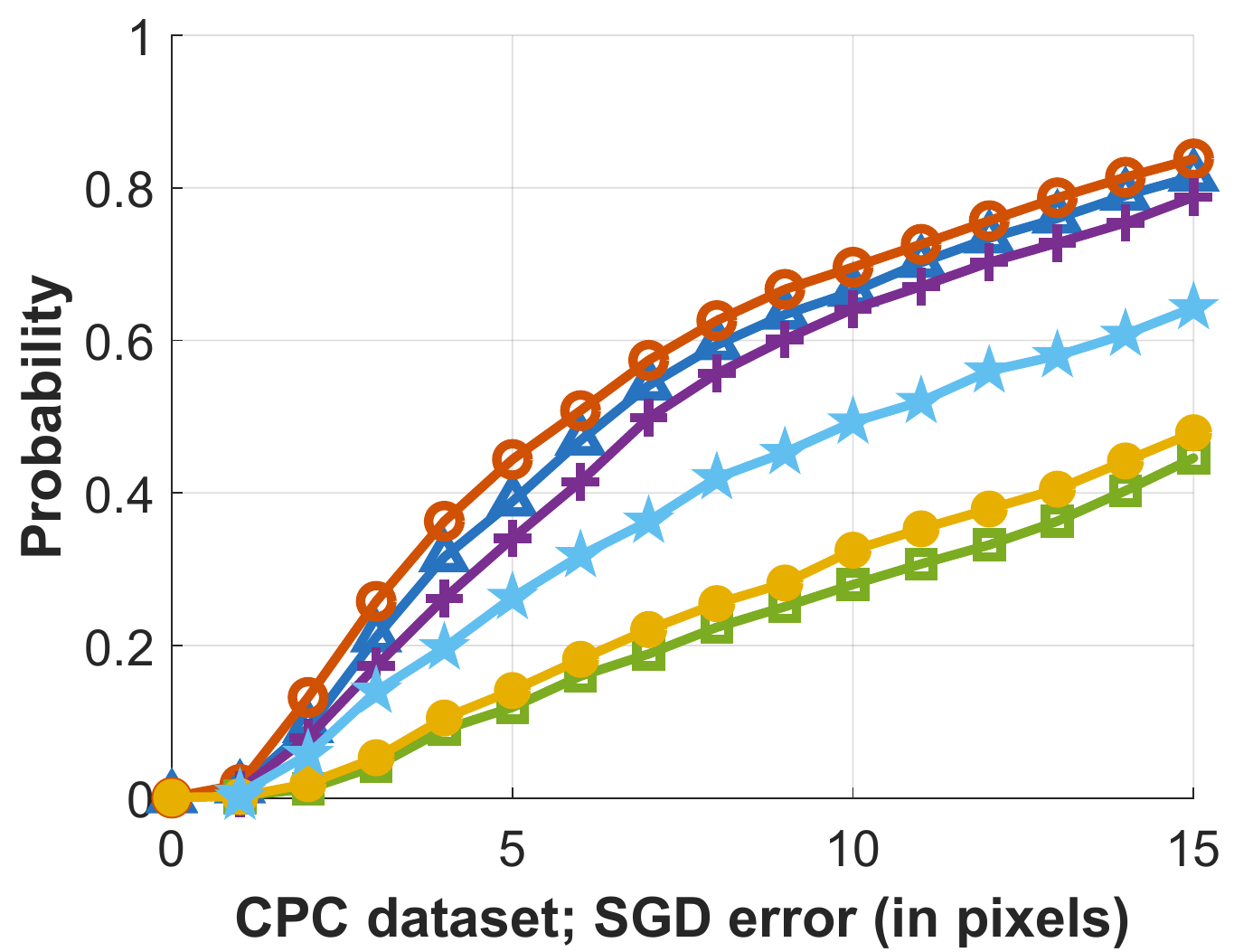}\hfill
    \includegraphics[width=0.495\columnwidth]{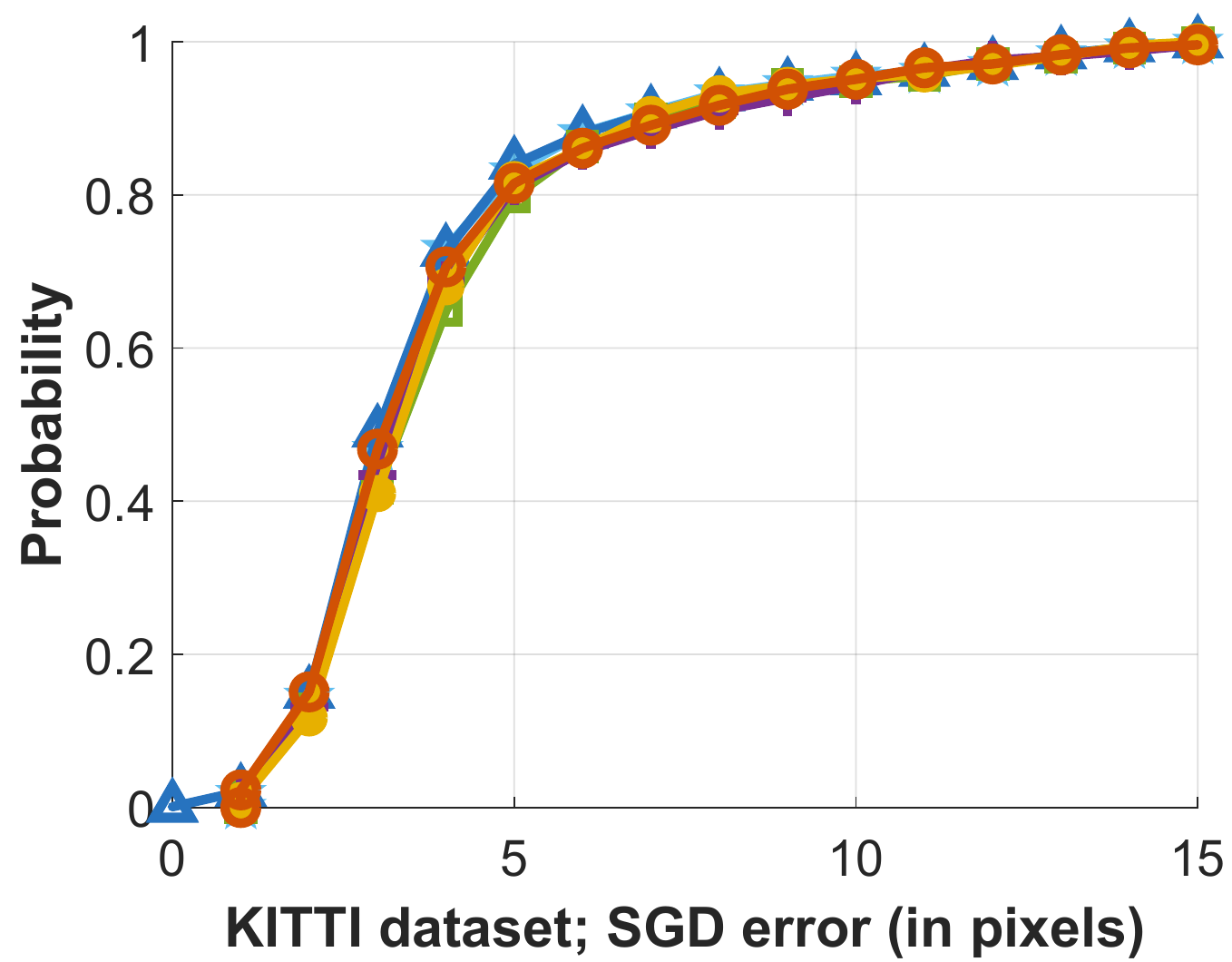}
    \caption{ The cumulative distribution functions (CDF) of the SGD errors (horizontal axis) of the estimated fundamental matrices, on datasets \dataset{CPC}, \dataset{T\&T}, \dataset{KITTI} and \dataset{TUM}. Being accurate is interpreted by a curve close to the top. }
    \label{fig:magsac_fundamental_matrix_experiments}
\end{figure}

In Fig.~\ref{fig:magsac_fundamental_matrix_experiments}, the cumulative distribution functions (CDF) of the SGD errors (horizontal) are shown. 
It can be seen that \textit{MAGSAC++ is the most accurate robust estimator} on \dataset{CPC}, \dataset{Tanks and Temples} and \dataset{TUM} datasets since its curve is always higher than that of the other methods. 
In \dataset{KITTI}, the image pairs are subsequent frames of a camera mounted to a car, thus, having short baseline. These image pairs are therefore easy and all methods lead to similar accuracy. 

In Table~\ref{tab:magsac_real_experiments}, the median errors (in pixels), the failure rates (in percentage) and processing times (in milliseconds) are reported.
We report the median values to avoid being affected by the failures -- which are also shown. 
A test is considered failure if the error of the estimated model is bigger than the $1\%$ of the image diagonal.
The best values are shown in red, the second best ones are in blue. 
It can be seen that for fundamental matrix fitting (first four datasets), \textit{MAGSAC++ is the best method on three datasets} both in terms of median error and failure rate.
On \dataset{KITTI}, all methods have similar accuracy -- the difference between the accuracy of the least and most accurate ones is $0.3$ pixel.
There, MAGSAC++ is the fastest.
On the tested datasets, MAGSAC++ is usually as fast as other robust estimators while leading to superior accuracy and failure rate. 
MAGSAC++ is \textit{always faster} than MAGSAC, e.g.,\ on \dataset{KITTI} by two orders of magnitude. 

\begin{figure}[t]
    \centering
    \phantom{xxx}\includegraphics[width=0.90\columnwidth]{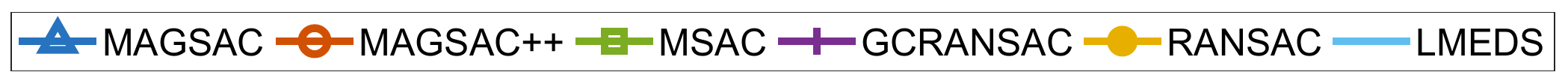}\\
	\begin{subfigure}[t]{0.49\columnwidth}
        \includegraphics[width=1.0\columnwidth]{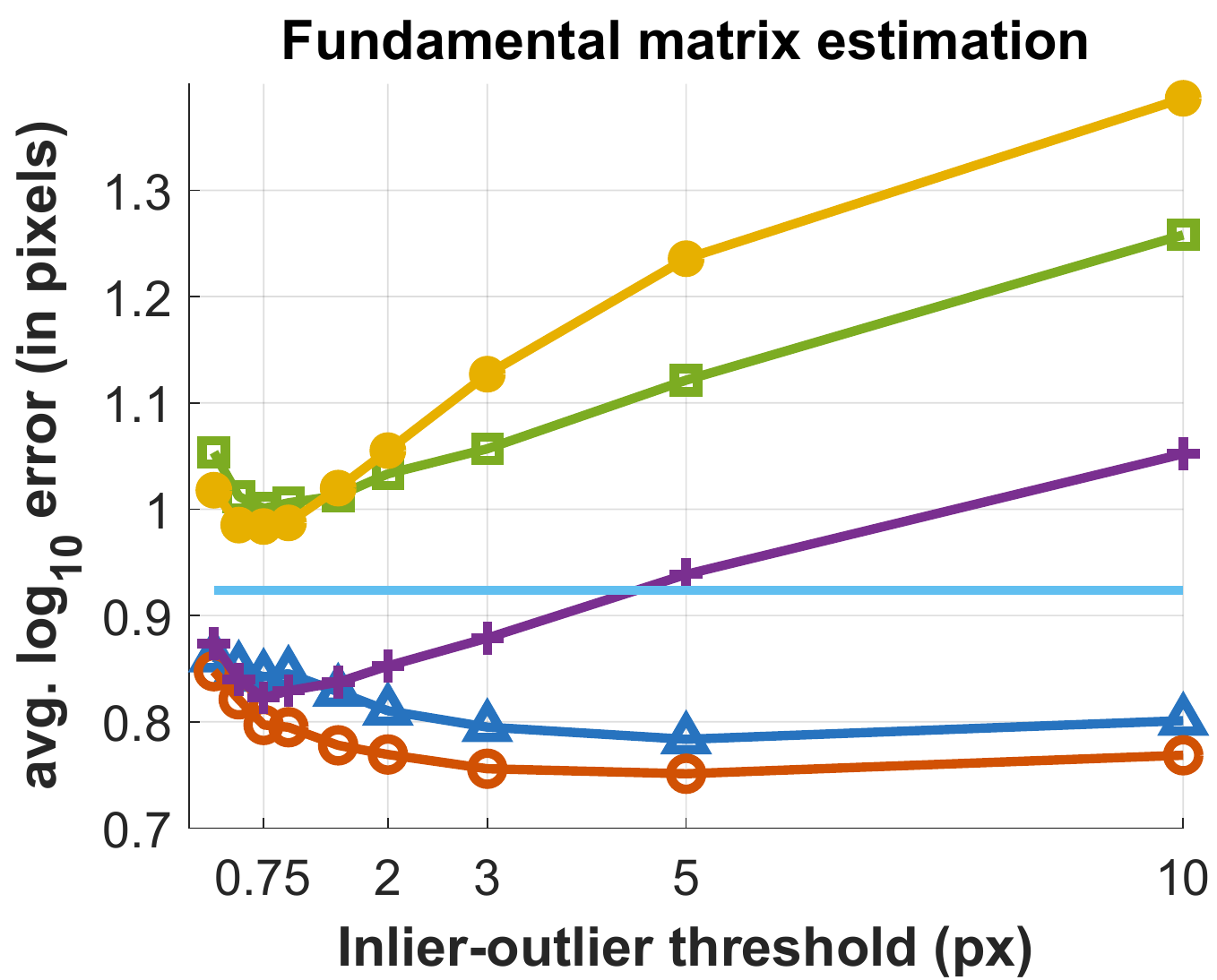}
    \end{subfigure}
    \hfill
	\begin{subfigure}[t]{0.49\columnwidth}
        \includegraphics[width=1.0\columnwidth]{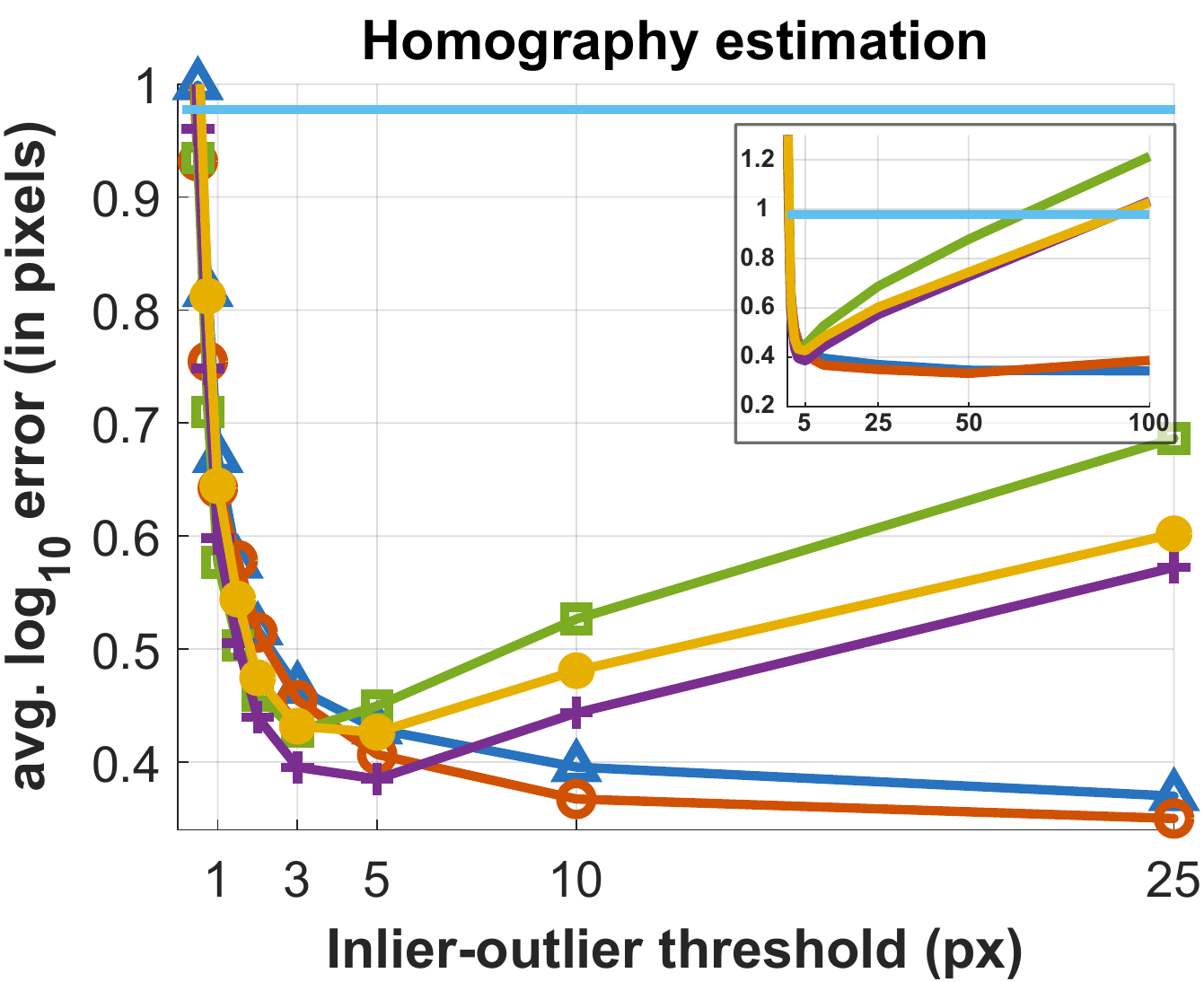}
    \end{subfigure}
    \caption{ The average $\log_{10}$ errors on the datasets for fundamental matrix (left; SGD error) and homography (right; RMSE re-projection error) fitting plotted as the function of the inlier-outlier threshold (in pixels). In the small plot inside the right one, the threshold goes up to $100$ pixels. }
    \label{fig:magsac_threshold_experiments}
\end{figure}

In the left plot of Fig.~\ref{fig:magsac_threshold_experiments}, the avg.\ $\log_{10}$ errors over all datasets are plotted as the function of the inlier-outlier threshold. 
It can be seen that both \textit{MAGSAC and MAGSAC++ are significantly less sensitive to the threshold} than the other robust estimators. 
Note that the accuracy of LMeDS is the same for all threshold values since it does not require an inlier-outlier threshold to be set. 

\begin{figure}[t]
    \centering
    \includegraphics[width=0.495\columnwidth]{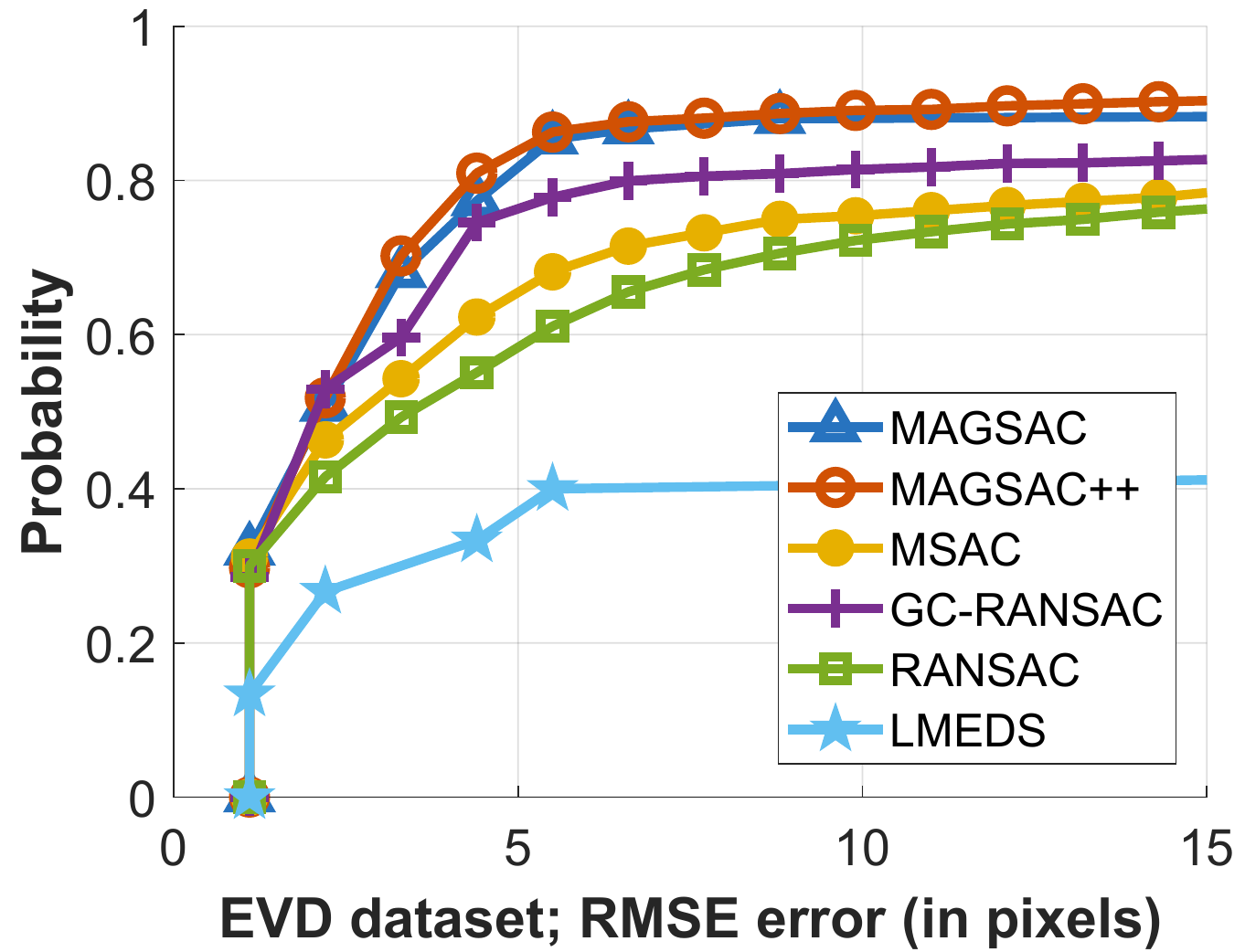}\hfill
    \includegraphics[width=0.495\columnwidth]{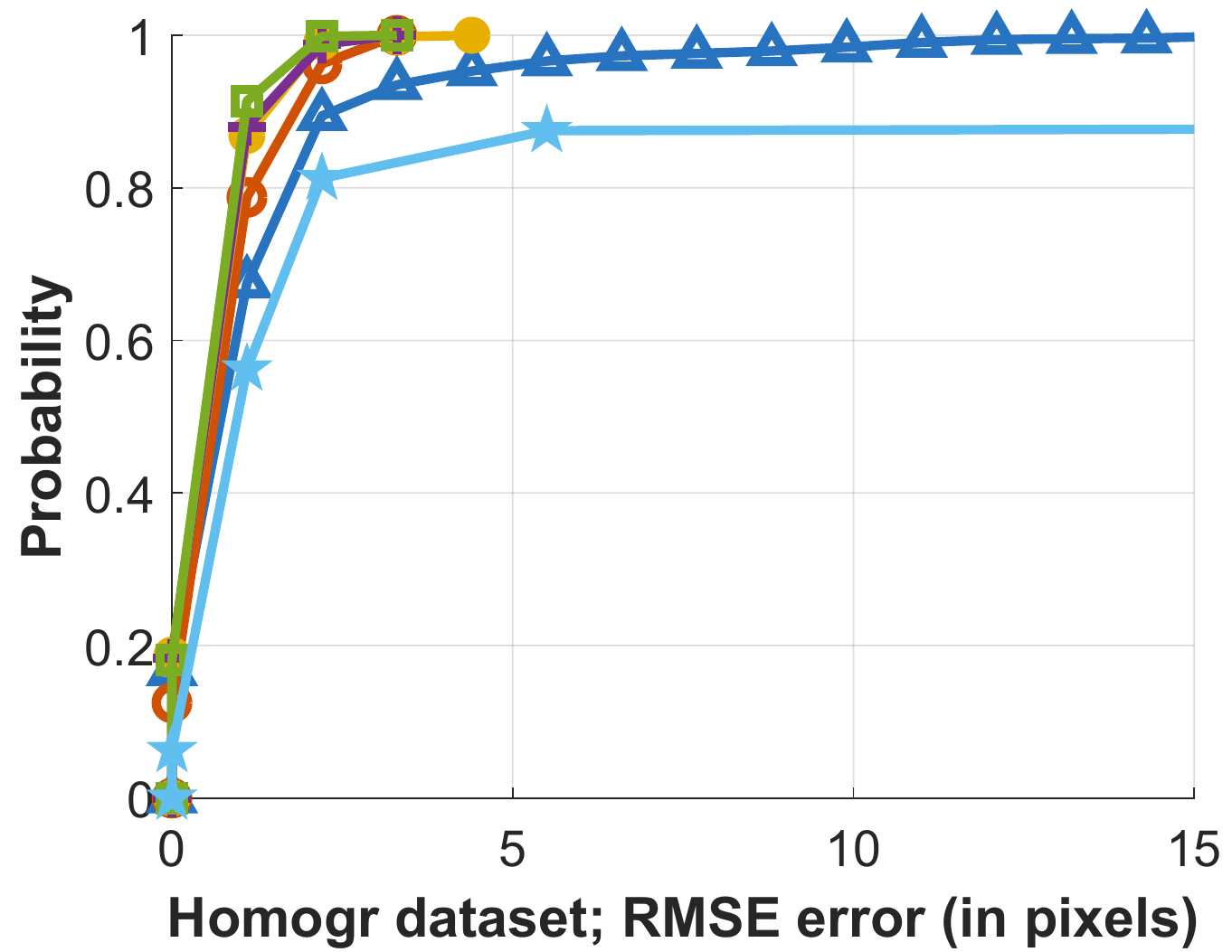}
    \caption{ The cumulative distribution functions (CDF) of the RMSE re-projection errors (horizontal axis) of the estimated homographies on datasets \dataset{EVD} and \dataset{homogr}. Being accurate is interpreted by a curve close to the top.}
    \label{fig:magsac_homography_experiments}
\end{figure}

\begin{table*}[t!]
	\centering
	\resizebox{1.00\linewidth}{!}{\begin{tabular}{ l c c c | c c c | c c c | c c c | c c c | c c c }
		\Xhline{2\arrayrulewidth}
		& \multicolumn{12}{c}{Fundamental matrix (Fig.~\ref{fig:magsac_fundamental_matrix_experiments})} & \multicolumn{6}{c}{Homography (Fig.~\ref{fig:magsac_homography_experiments})} \\
		\cdashlinelr{1-19}
		 & \multicolumn{3}{c}{\dataset{KITTI}~\cite{geiger2012we}} & \multicolumn{3}{c}{\dataset{TUM}~\cite{sturm2012benchmark}} & \multicolumn{3}{c}{\dataset{T\&T}~\cite{knapitsch2017tanks}} & \multicolumn{3}{c}{\dataset{CPC}~\cite{wilson2014robust}} & \multicolumn{3}{c}{\dataset{Homogr}~\cite{lebeda2012fixing}} & \multicolumn{3}{c}{\dataset{EVD}~\cite{lebeda2012fixing}} \\
		\hline
		 & $\epsilon_{\text{med}}$ & $\lambda$ & $t$ & $\epsilon_{\text{med}}$ & $\lambda$ & $t$ & $\epsilon_{\text{med}}$ & $\lambda$ & $t$ & $\epsilon_{\text{med}}$ & $\lambda$ & $t$ & $\epsilon_{\text{med}}$ & $\lambda$ & $t$ & $\epsilon_{\text{med}}$ & $\lambda$ & $t$ \\
		\hline 
		\multicolumn{1}{l |}{\Rowcolor{lightgray} MAGSAC++} & \second{3.6} & \second{2.4} & \phantom{11}\win{8} & \win{3.5} & \win{16.4} & 13 & \win{3.9} & \win{0.4} & 142 & \phantom{1}\win{6.4} & \phantom{1}\win{7.8} & 156 & \win{1.1} & \phantom{1}\win{0.0} & \phantom{1}\win{6} & \phantom{1}\win{2.6} & \win{10.4} &  173 \\
        \multicolumn{1}{l |}{MAGSAC} & \win{3.5} & 2.8 & 117 & \second{3.7} & \second{17.7} & 18 & \second{4.2} & \second{0.7} & 267 & \phantom{1}\second{7.0} & \phantom{1}\win{7.8} & 261 & \second{1.3} & \phantom{1}\second{0.8} & 32 & \phantom{1}\win{2.6} & \second{12.0} & 426 \\ 
        \multicolumn{1}{l |}{GC-RANSAC} & 3.7 & \win{2.3} & \phantom{1}11 & 4.1 & 25.1 & \win{11} & 4.5 & 2.2 & \win{126} & \phantom{1}7.5 & \second{12.1} & \win{144} & \win{1.1} & \phantom{1}\win{0.0} & 25 & \phantom{1}\win{2.6} & 18.3 & \phantom{1}\second{66} \\
        \multicolumn{1}{l |}{RANSAC} & 3.8 & 2.7 & \phantom{11}\second{9} & 5.4 & 22.1 & \win{11} & 6.3 & 2.6 & \second{133} & 16.9 & 29.5 & \second{151} & \win{1.1} & \phantom{1}\win{0.0} & 26 & \phantom{1}4.0 & 26.1 & \phantom{1}68 \\
        \multicolumn{1}{l |}{LMedS} & \second{3.6} & 2.7 & \phantom{1}11 & 4.3 & 23.9 & \second{12} & 4.9 & 1.1 & 166 & 10.7 & 17.8 & 187 & 1.5 & 12.5 &  31& 89.9 & 60.0 & \phantom{1}82 \\
        \multicolumn{1}{l |}{MSAC} & 3.8 & 2.6 & \phantom{1}10 & 5.5 & 36.2 & \win{11} & 7.0 & 2.2 & \second{133} & 16.5 & 33.8 & 153 & \win{1.1} & \phantom{1}\win{0.0} &  \second{24}& \phantom{1}\second{3.2} & 23.7 & \phantom{1}\win{64} \\
		\Xhline{2\arrayrulewidth}
	\end{tabular}}
	\caption{The median errors ($\epsilon_{\text{med}}$; in pixels), failure rates ($\lambda$; in percentage) and average processing times ($t$, in milliseconds) are reported for each method (rows from 4th to 9th) on all tested problems (1st row) and datasets (2nd). 
	The error of fundamental matrices is calculated from the ground truth matrix as the symmetric geometric distance~\cite{zhang1998determining} (SGD). 
	For homographies, it is the RMSE re-projection error from ground truth inliers.
	A test is considered failure if the error is bigger than the $1\%$ of the image diagonal.
	For each method, the inlier-outlier threshold was set to maximize the accuracy and the confidence to $0.99$. 
	The best values in each column are shown by red and the second best ones by blue. Note that all methods, excluding MAGSAC and MAGSAC++, finished with a final LS fitting on all inliers. }
    \label{tab:magsac_real_experiments}
\end{table*}

\customparagraph{For homography} estimation, we downloaded \dataset{homogr} (16 pairs) and \dataset{EVD} (15 pairs) datasets~\cite{lebeda2012fixing}. 
They consist of image pairs of different sizes from $329 \times 278$ up to $1712 \times 1712$ with point correspondences and inliers selected manually.   
The \dataset{homogr} dataset contains mostly short baseline stereo images, whilst the pairs of \dataset{EVD} undergo an extreme view change, i.e., wide baseline or extreme zoom. 
In both datasets, the correspondences are assigned manually to one of the two classes, i.e., outlier or inlier of the most dominant homography present in the scene. 
All algorithms applied the normalized four-point algorithm~\cite{hartley2003multiple} for homography estimation and were repeated $100$ times on each image pair. 
To measure the quality of the estimated homographies, we used the RMSE re-projection error calculated from the provided ground truth inliers. 

The CDFs of the errors are shown in Fig.~\ref{fig:magsac_homography_experiments}.
It can be seen that, on \dataset{EVD}, the MAGSAC++ goes the highest -- it is the most accurate method. 
On \dataset{homogr}, all methods but the original MAGSAC and LMedS have similar accuracy. 
The last two datasets in Table~\ref{tab:magsac_real_experiments} report the median errors, failure rates and runtimes.
It can be seen that, on \dataset{EVD}, \textit{MAGSAC++ failed the least often, while having the best median accuracy and being $2.5$ times faster} than MAGSAC. 
All of the faster methods fail to return the sought model significantly more often. 
On \dataset{homogr}, MAGSAC++, GC-RANSAC, RANSAC and MSAC have similar results.
MAGSAC++ is the fastest one amongst them by almost an order of magnitude.

In the right plot of Fig.~\ref{fig:magsac_threshold_experiments}, the avg.\ $\log_{10}$ errors are plotted as the function of the inlier-outlier threshold (in px). 
Both \textit{MAGSAC and MAGSAC++ are significantly less sensitive to the threshold} than the other robust estimators. 
In the small figure, inside the bigger one, the threshold value goes up to $100$ pixels.
For MAGSAC and MAGSAC++, parameter $\sigma_{\text{max}}$ was calculated from the threshold value.

In summary, the experiments showed that MAGSAC++ is more accurate on the tested problems and datasets than all the compared state-of-the-art robust estimators with being significantly faster than the original MAGSAC.


\subsection{Evaluating Progressive NAPSAC}

In this section, the proposed P-NAPSAC sampler is evaluated on homography and fundamental matrix fitting using the same datasets as in the previous sections. 
Every tested sampler is combined with MAGSAC++. 

The compared samplers are the uniform sampler of plain RANSAC~\cite{fischler1981random}, NAPSAC~\cite{nasuto2002napsac}, PROSAC~\cite{chum2005matching},
and the proposed P-NAPSAC.
Since both the proposed P-NAPSAC and NAPSAC assumes the inliers to be localized, they used the relaxed termination criterion with $\gamma = 0.1$. Thus, they terminate when the probability of finding a model which leads to at least $0.1$ increment in the inlier ratio falls below a threshold. 
PROSAC used its original termination criterion and the quality function for sorting the correspondences was the one proposed in the original paper~\cite{chum2005matching}.

Example image pairs are shown in Fig.~\ref{fig:example_result_2}.
Inlier correspondences are marked by line segments joining the corresponding points. 
The numbers of iterations and processing times of PROSAC or P-NAPSAC samplers are reported in the captions. 
In both cases, P-NAPSAC leads to significantly fewer iterations than PROSAC. 
\begin{figure}[t]
    \centering
	\begin{subfigure}[t]{0.520\columnwidth}
     \includegraphics[width=1.0\columnwidth]{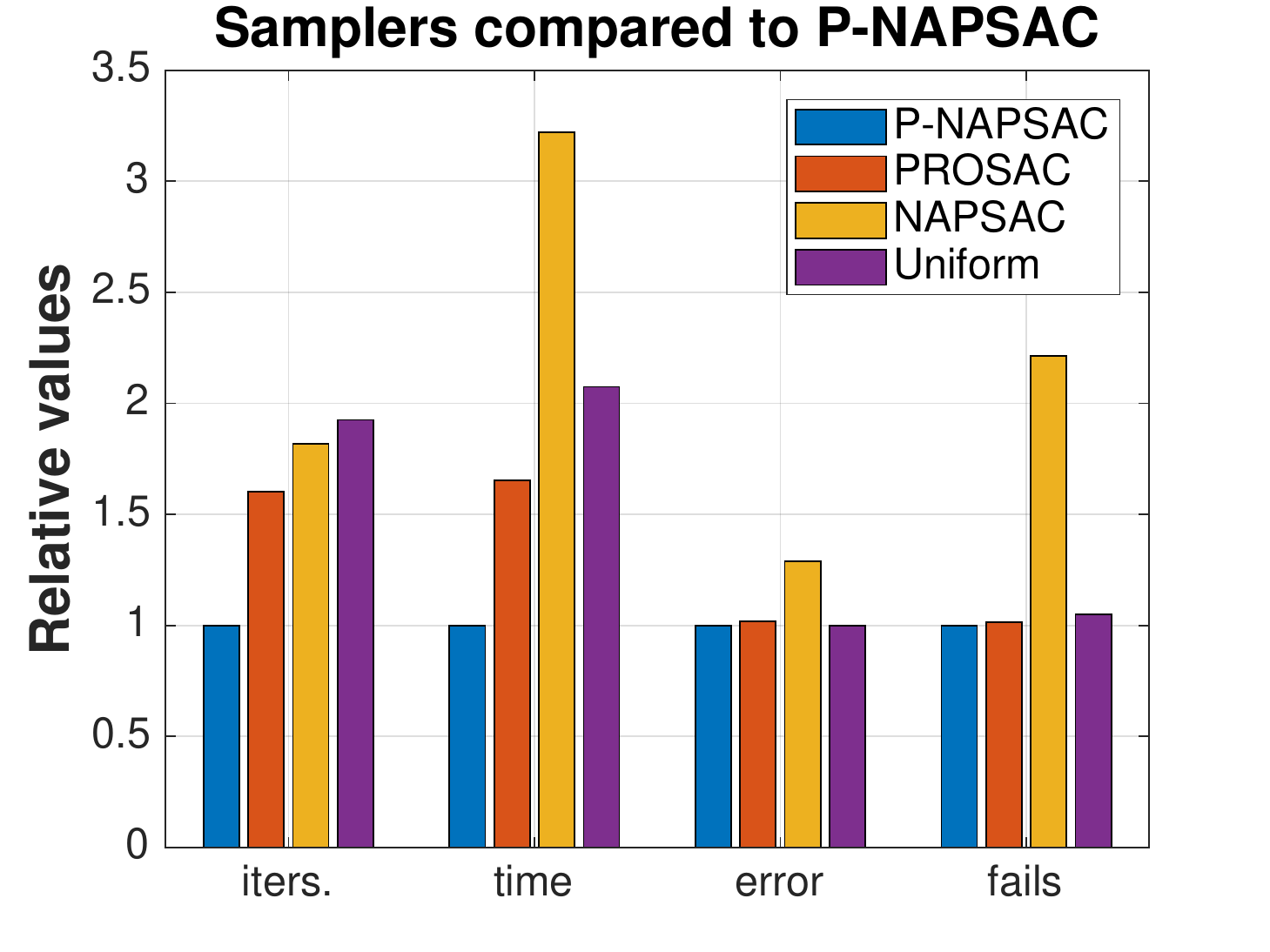}
     \caption{\label{fig:pnapsac_real_experiments}}
	\end{subfigure}
	\begin{subfigure}[t]{0.47\columnwidth}
    	\includegraphics[width=1.0\columnwidth]{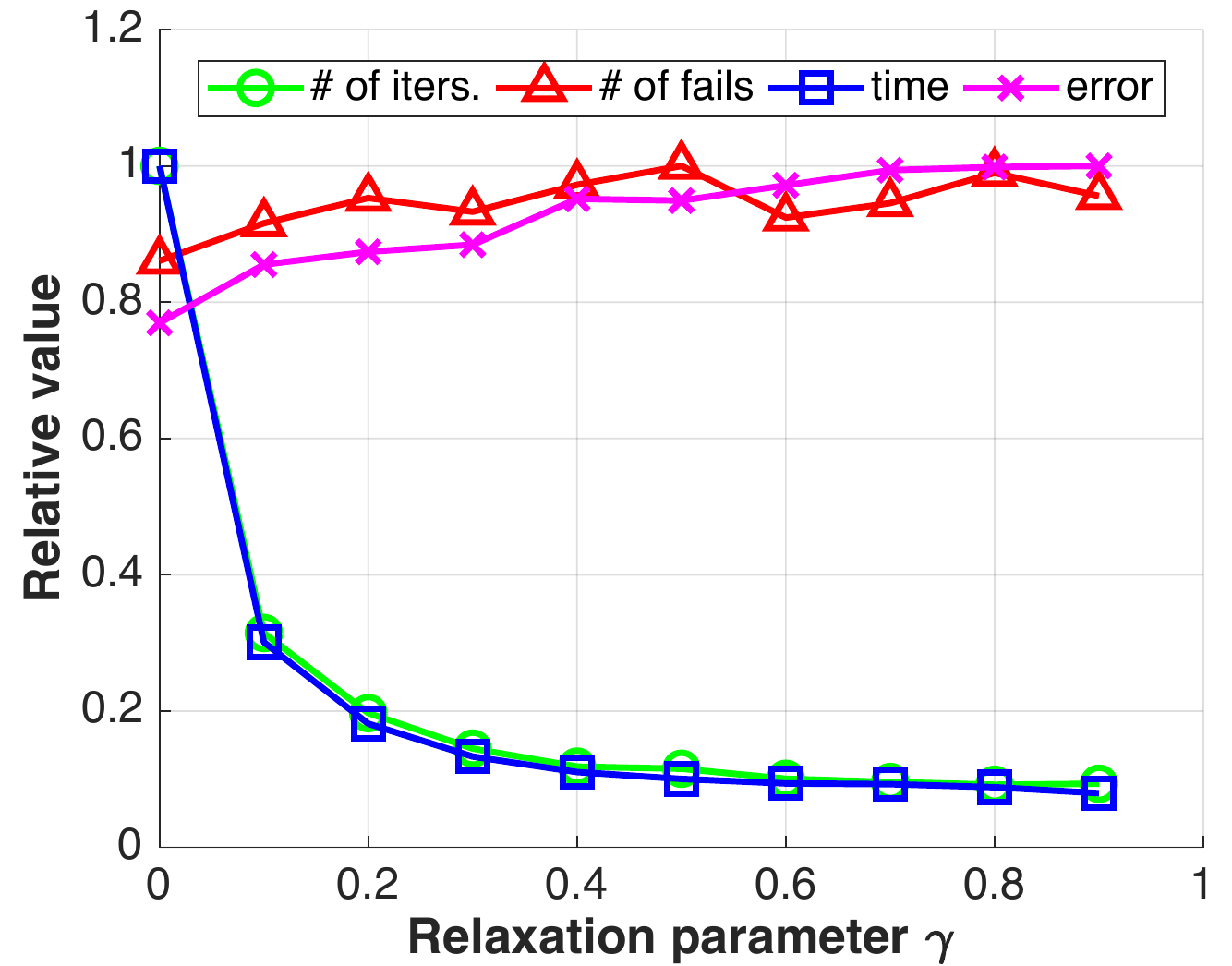}
     \caption{\label{fig:relaxation}}
	\end{subfigure}
    \caption{\textbf{(a)} Comparison of samplers to P-NAPSAC (blue bar; divided by the values of P-NAPSAC; all combined with MAGSAC++) on the datasets of Table~\ref{tab:magsac_real_experiments}. The reported properties are: the \# of iterations; processing time; average error; and failure rate.
    \textbf{(b)} The relative (i.e., divided by the maximum) error, number of fails, processing time, and number of iterations are plotted as the function of the relaxation parameter $\gamma$ (from Eq.~\ref{eq:relaxed_ransac_iterations}) of the relaxed RANSAC termination criterion.}
    
\end{figure}
The results of the samplers, compared to P-NAPSAC and averaged over all datasets, are shown in Fig.~\ref{fig:pnapsac_real_experiments}.
The number of iterations and, thus, the processing time is the lowest for P-NAPSAC. 
It is approx.\ $1.6$ times faster than the second best, i.e., PROSAC, while being similarly accurate with the same failure rate.

\customparagraph{Relaxed termination criterion.}
To test the relaxed criterion, we applied P-NAPSAC to all datasets using different $\gamma$ values. 
We then measured how each property (i.e., the error of the estimated model, failure rate, processing time, and number of iterations) changes.
Fig.~\ref{fig:relaxation} plots the average (over $100$ runs on each scene) of the reported properties as the function of $\gamma$. 
The relative values are shown. Thus, for each test, the values are divided by the maximum. 
For instance, if P-NAPSAC draws $100$ iterations when $\gamma = 0$, the number of iterations is divided by $100$ for every other $\gamma$. 

It can be seen that the error and failure ratio slowly increase from approximately $0.8$ to $1.0$. The trend seems to be close to linear. 
Simultaneously, the number of iterations and, thus, the processing time decrease dramatically.
Around $\gamma = 0.1$ there is significant drop from $1.0$ to $0.3$. 
If $\gamma > 0.1$ both values decrease mildly.
Therefore, selecting $\gamma = 0.1$ as the relaxation factor does not lead to noticeably worse results but speeds up the procedure significantly. 

\section{Conclusion}

In the paper, two contributions were made. First, 
we formulate a novel marginalization procedure as an iteratively re-weighted least-squares approach 
and we introduce a new model quality (scoring) function that does not require the inlier-outlier decision.
Second, we propose a new sampler, Progressive NAPSAC, for RANSAC-like robust estimators. 
Reflecting the fact that nearby points often originate from the same model in real-world data, P-NAPSAC finds local structures earlier than global samplers.
The progressive transition from local to global sampling does not suffer from the weaknesses of purely localized samplers.

The two orthogonal improvements are combined with the "bells and whistles" of USAC~\cite{raguram2013usac}, e.g., pre-emptive verification, degeneracy testing.
On six publicly available real-world datasets for homography and fundamental matrix fitting,
MAGSAC++ produces results superior to the state-of-the-art robust methods. 
It is faster, more geometrically accurate and fails less often.

{\small
\bibliographystyle{ieee}
\bibliography{egbib}
}

\end{document}